\documentclass[runningheads]{llncs}

\usepackage[final]{eccv}
\usepackage{eccvabbrv}
\usepackage{overpic}
\usepackage{amsmath}
\usepackage{amssymb}
\usepackage{xcolor}
\usepackage{pgfplots}
\usepackage{pgfplotstable}
\pgfplotsset{compat=newest}
\usepackage{algorithm}
\usepackage{algpseudocode}
\algrenewcommand\algorithmicrequire{\textbf{Input:}}
\algrenewcommand\algorithmicensure{\textbf{Output:}}
\usepackage{float}
\usepackage{multicol,multirow}
\usepackage{graphicx}
\usepackage{booktabs}
\usepackage{colortbl}
\definecolor{action2}{RGB}{130,179,102} 
\definecolor{action3}{RGB}{150,115,166} %
\usepackage{soul}
\usepackage[accsupp]{axessibility}  %
\usepackage{hyperref}
\usepackage{orcidlink}
\DeclareMathOperator*{\argmax}{arg\,max}

\begin{document}

\title{Adaptive Latent Trajectory Anchoring for\\ Action Segmentation Dataset Condensation} 

\titlerunning{Adaptive Latent Trajectory Anchoring}

\author{Arthème Gauthier-Villars\inst{1,2}$^{*}$\orcidlink{0009-0004-5771-8088} \and
Guodong Ding\inst{1}$^{*,\dagger}$\orcidlink{0000-0001-6080-5220} \and
Angela Yao\inst{1}\orcidlink{0000-0001-7418-6141}}
\authorrunning{A. Gauthier-Villars, G. Ding and A. Yao}
\institute{National University of Singapore, Singapore \and
ETH Zurich, Switzerland\\
\email{agauthier@ethz.ch\quad \{dinggd,ayao\}@comp.nus.edu.sg}}

\maketitle
\begingroup
\renewcommand{\thefootnote}{}
\footnotetext{$^{*}$ Equal contribution.}
\footnotetext{$^{\dagger}$ Corresponding author and project lead.}
\endgroup

\begin{abstract}
    Dataset condensation for action segmentation synthesizes compact, informative representations of long, untrimmed video datasets. The existing approach relies on Variational Autoencoders and an iterative latent optimization; it is computationally expensive and suffers from over-smoothed reconstructions and rigid temporal constraints. 
    This paper proposes to shift the condensation paradigm from optimization-based inversion to deterministic latent mapping. By leveraging Denoising Diffusion Implicit Models, we represent action segments as continuous trajectories anchored by sparse latent points in the noise manifold. 
    To maximize representational efficiency, we introduce an adaptive allocation mechanism that dynamically redistributes the anchoring budget based on segment-wise reconstruction difficulty.
    Extensive experiments demonstrate that our framework significantly outperforms state-of-the-art methods in segmentation performance across common datasets. Notably, our approach achieves performance parity with real data training while maintaining a condensation ratio of 2.4\% on Breakfast dataset.
  \keywords{Dataset Condensation \and Temporal Action Segmentation \and Diffusion Models}
\end{abstract}

\begin{figure*}[t]         %
  \centering
\begin{overpic}[width=\linewidth, grid=false]{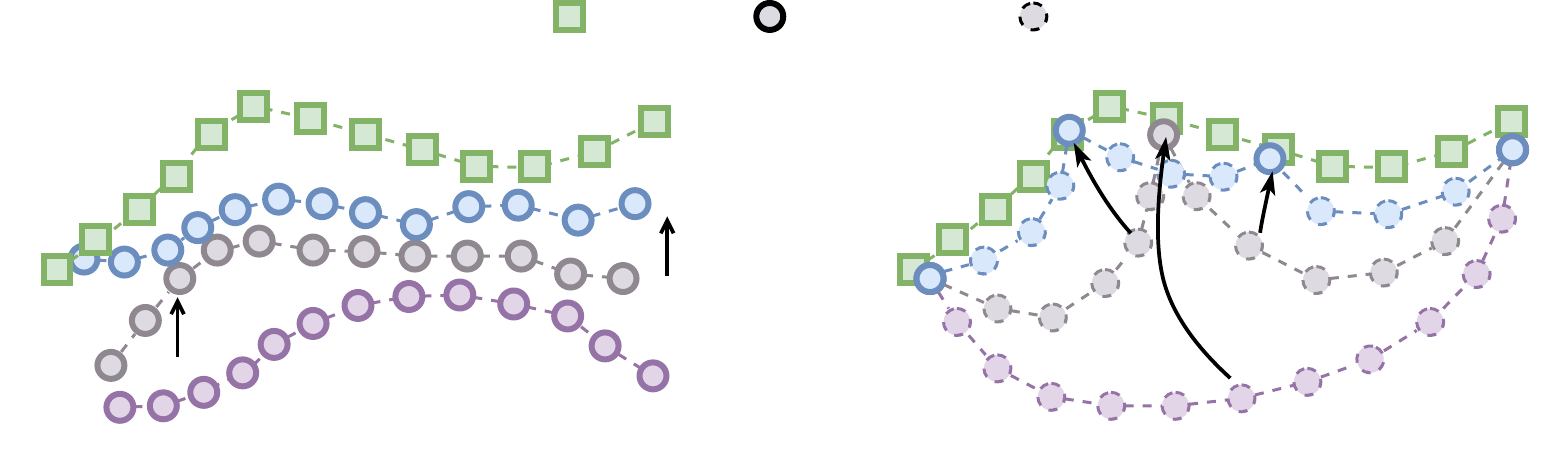}
    \put(43.5,13){\tiny $O_{k}$}
    \put(11.8,9){\tiny $O_{k-1}$}
    \put(68,16){\tiny $A_2$}
    \put(75,11){\tiny $A_1$}
    \put(81,16){\tiny $A_2$}
    \put(38,28.5){\tiny Original}
    \put(51,28.5){\tiny Reconstructed}
    \put(68.5,28.5){\tiny Interpolated}
    \put(11,0){\scriptsize (a) Direct Optimization~\cite{ding2025condensing}}
    \put(63,0){\scriptsize (b) Latent Anchoring (Ours)}
\end{overpic}
  \caption{
Comparison of TAS condensation paradigms. (a) Direct Optimization: An existing method uses iterative optimizations ($O_k, O_{k-1}$) to find optimal latent codes for condensation and rely on these fixed codes for reconstruction, which can cause low reconstruction fidelity. (b) Latent Anchoring: Our method adaptively selects latent anchors ($A_1,A_2$) per segment for condensation and uses latent trajectory interpolation to enable reconstruction of fine grained action dynamics.  
}\label{fig:teaser}
\end{figure*}
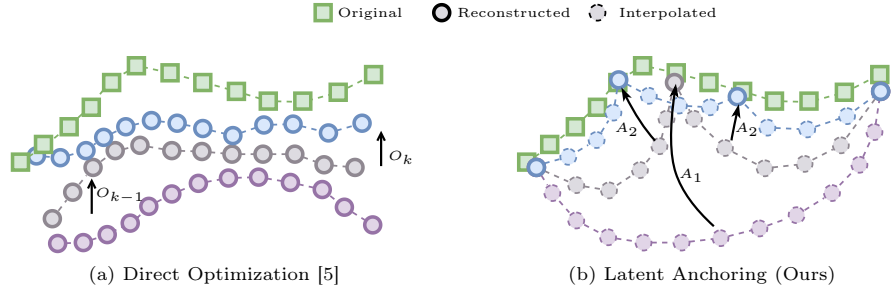

\section{Introduction}
Temporal action segmentation (TAS)~\cite{ding2023temporal} targets the task of predicting a semantic label for every frame in a long, untrimmed video.  Despite the success of modern TAS architectures~\cite{farha2019ms,yi2021asformer,liu2023diffusion}, their performance remains heavily dependent on the availability of massive, densely-labeled datasets, which introduces significant bottlenecks in terms of storage and training efficiency. To address this, dataset condensation~\cite{wang2018dataset} has emerged as a promising direction, seeking to synthesize a compact, highly informative representation of the original data.

A recent pioneering study formulates this problem as generative network inversion~\cite{ding2025condensing}. A conditional Variational Autoencoder (cVAE)~\cite{kingma2014auto} is first trained to model action priors in the feature space and each action segment is then reconstructed by optimizing latent variables to minimize reconstruction error. While effective, this strategy inherits structural constraints: {reconstruction fidelity} depends on the expressiveness of a variational latent space trained under a unimodal Gaussian prior, and condensation requires iterative per-segment optimization. As a result, {fine temporal variations} may be over-smoothed. More importantly, the same latent code is reused across consecutive frames during reconstruction, imposing a block-wise temporal structure that may even obscure the continuous evolution of action dynamics.

In this work, we depart from the inversion perspective and revisit TAS condensation through the lens of generative dynamics. Our central observation is that action segments are not arbitrary collections of feature vectors; they exhibit structured, smooth trajectories in representation space. A desirable condensation mechanism should therefore preserve this trajectory structure rather than compress segments into isolated random codes. This requirement calls for a generative model whose latent space admits a deterministic and structure-preserving mapping between data and noise.  Deterministic diffusion models, in particular DDIMs~\cite{song2021ddim}, are a natural fit. Unlike cVAEs, DDIM defines a deterministic reverse process that establishes an almost bijective mapping between a data sample and its corresponding noise. As a result, any input can be mapped to its corresponding latent, enabling high-fidelity reconstruction without the need for per-instance optimization. 

Building on this property, we propose a diffusion-based TAS condensation framework that reformulates compression as a latent trajectory anchoring problem. We represent action segments as  continuous trajectories in diffusion latent space, anchored by sparse, informative points. 
Specifically, each action segment is encoded into the latent trajectory induced by the deterministic DDIM reverse process. To achieve condensation, we store each segment as a sparse collection of latent anchors that characterizes the trajectory.  During reconstruction, the full feature sequence can be recovered by interpolating between latent anchors. 

Going a step further, we introduce an adaptive anchoring strategy that allocates anchors according to segment-specific reconstruction difficulty by dynamically adjusting anchor density along the trajectory. In particular, segments exhibiting more complex temporal variations are assigned higher anchor density to better preserve reconstruction fidelity, while simpler segments are represented using fewer anchors for better efficiency.

Our proposed framework offers several advantages. First, by leveraging deterministic diffusion trajectories, the condensation process avoids expensive per-segment latent optimization while still supporting faithful sequence reconstruction. Second, interpolation between the latent anchors helps reconstruct fine-grained temporal variations as oppose to  decoding from repeated latent codes. Third, the adaptive anchoring strategy further improves efficiency by allocating higher anchor density to segments with higher reconstruction difficulty. 
A conceptual comparison between the existing optimization-based paradigm~\cite{ding2025condensing} and our proposed latent anchoring framework is provided in~\cref{fig:teaser}.

\noindent\textbf{Contributions.} Our contributions are summarized as follows: 1) We offer a novel perspective for TAS condensation by recasting the problem as a generative latent trajectory reconstruction task through the lens of diffusion models. This shift enables high-fidelity feature synthesis while bypassing iterative inversion. 2) We propose latent trajectory anchoring with latent-space interpolation as a replacement for discrete code instantiation, modeling action segments as continuous trajectories that preserve fine-grained temporal dynamics. 3) We develop an adaptive budgeting strategy that redistributes representation capacity based on reconstruction difficulty, ensuring scalability and flexibility across diverse action lengths. 4) We demonstrate that our framework achieves significant dataset condensation, with state-of-the-art performance across major benchmarks.

\section{Related Work}
\noindent\textbf{Temporal Action Segmentation.} TAS requires the frame-level labeling of action sequences in untrimmed videos~\cite{ding2023temporal}. The field primarily focuses on architecture, supervision, and training paradigms. Architectural designs prioritize long-range temporal modeling, spanning dilated convolutional networks (MSTCN~\cite{farha2019ms}), transformer-based architectures (ASFormer~\cite{yi2021asformer}), and diffusion-based generative frameworks (DiffAct~\cite{liu2023diffusion}). To mitigate annotation costs, diverse supervision regimes have been studied ranging from fully supervised~\cite{farha2019ms,yi2021asformer,liu2023diffusion,singhania2023c2f, lu2024fact}, semi-supervised~\cite{ding2022leveraging, singhania2022iterative} to weakly supervised~\cite{richard2018action, ding2022temporal} and unsupervised learning~\cite{kumar2022unsupervised, bueno2025clot}. Furthermore, the scope extends to specialized settings including active learning~\cite{su2024two}, incremental learning \cite{ding2024coherent}, and online learning~\cite{shen2024progress, zhong2024onlinetas}. Despite this breadth, the challenge of TAS condensation~\cite{ding2025condensing} remains an emerging frontier. Unlike previous work, we replace the optimization-based inversion paradigm with deterministic latent mapping to condense action segments into sparse latent anchors and restore through latent trajectory interpolation.

\noindent\textbf{Dataset Condensation.}
DC aims to synthesize a small set of informative samples that represent the knowledge of a large-scale dataset~\cite{wang2018dataset}. While initial methods focused on image classification~\cite{cazenavette2022dataset,cui2023scaling,liu2022dataset,wang2022cafe,zhao2023dataset} using gradient matching, trajectory matching~\cite{cazenavette2022dataset} or distribution matching~\cite{zhao2023dataset}. Extending DC to the video domain is challenging due to the high dimensionality and temporal redundancy of video features. Effective condensation must therefore preserve temporal coherence and transition structure and directly adapting image-based strategies is thus suboptimal. Furthermore, TAS represents a structured video setting that further requires fine-grained, frame-wise temporal modeling, making action recognition condensation techniques~\cite{wang2024dancing} undesirable. Existing TAS condensation methods~\cite{ding2025condensing} rely on iterative inversion of generative models like cVAEs. However, these are often computationally expensive and struggle to capture continuous temporal dynamics. In contrast, our approach leverages the deterministic flow of DDIM to achieve efficient, trajectory-based condensation without the need for costly per-sequence optimization. 

\section{Preliminaries}

\subsection{Temporal Action Segmentation}

Temporal Action Segmentation (TAS)~\cite{ding2023temporal} aims to assign a semantic category to every frame in an untrimmed video. 
Formally, a video is represented as a sequence of $L$ frame-level features $V=\{x_i\}_{i=1}^L$, where $x_i\in \mathbb{R}^D$ represents the feature vector of the $i$-th frame from a pretrained visual backbone, \eg, I3D~\cite{carreira2017quo}. The goal is to map this sequence to a corresponding set of semantic labels $Y=\{y_i\}_{i=1}^L$, where each $y_i\in[1, \dots, A]$ denotes the action class.

To learn a TAS model $\mathcal{M}$~\cite{farha2019ms,yi2021asformer}, a composite objective function is typically employed to balance the frame-wise accuracy with temporal coherence. The primary component is the classification loss, formulated as a frame-wise cross-entropy:
\begin{equation}
    \mathcal{L}_\text{cls} (x,y) = \frac{1}{L}\sum_{i=1}^L-\log(\hat{y}_{i,a}),
\end{equation}
where $\hat{y}_{i,a}$ represents the predicted probability for the ground-truth class $a$ for frame $i$. In addition, to address the over-segmentation issue where the model predicts frequent and erratic action transition, a smoothing loss is integrated:
\begin{equation}\label{eq:sm}
    \mathcal{L}_{\text{sm}}(x)\! =\! \frac{1}{LA}\sum_{i,a}\tilde{\Delta}_{i,a}^2, \;
    \tilde{\Delta}_{i,a}\! =\! \begin{cases}
    \Delta_{i,a}\kern-0.8em &:\! \Delta_{i,a} \le \tau\\
    \tau \kern-0.8em&:\! \text{otherwise}
    \end{cases}, \Delta_{i,a}\! = \!\left|\log(\hat{y}_{i,a})\! -\! \log(\hat{y}_{i-1,a})\right|,
\end{equation}
with the truncation parameter $\tau$ set to 4, following~\cite{farha2019ms}. 
By minimizing the total loss with a trade--off parameter $\lambda$:
\begin{equation}
    \mathcal{L}_\text{tas} = \mathcal{L}_\text{cls} (x,y)+ \lambda \mathcal{L}_\text{sm} (x), \label{eq:tas}
\end{equation}
the model is encouraged to produce semantically accurate yet temporally stable segmentation. Complementing the frame-wise view, the segmentation can also be represented homogeneously as a sequence of $N$ action segments, \ie,:
\begin{equation}
    S=\{s_1, \dots, s_N\}, \quad \text{where} \quad s_n = (a_n, t_n, \ell_n) \quad \text{and} \quad t_{n+1} = t_n+\ell_n,
\end{equation} 
where each segment $s_n$ is characterized by its action category $a_n$, its starting timestamp $t_n$, and its temporal duration $\ell_n$. 
\subsection{TAS Condensation}
The main objective of TAS condensation is to construct a compressed representation of a TAS dataset while preserving its task-relevant and temporal properties. The condensed dataset should retain sufficient semantic and sequential information such that a TAS model trained on it achieves performance comparable to that obtained when trained on the full, uncompressed dataset. 

Formally, given an original TAS dataset $\mathcal{D} = \{(V_i, Y_i)\}_{i=1}^{N_v}$, 
the goal of TAS condensation is to construct a compressed dataset $\mathcal{D}^* = \{(V^*_i, L_i)\}_{i=1}^{N_v}$ that enables the generation of a reconstructed proxy dataset $\hat{\mathcal{D}} = \{(\hat{V}_i, L_i)\}_{i=1}^{N_v}$
such that a model trained on $\hat{\mathcal{D}}$ achieves performance close to that obtained when trained on the original dataset $\mathcal{D}$. 
Each compressed representation is defined as $V^*_i = \{v^*_{i,1}, \dots, v^*_{i,T^*_i}\}$, with $v^*_{i,j} \in \mathbb{R}^{D}$ and $T^*_i\ll T_i$. We also follow~\cite{ding2025condensing} and define the compression ratio as $\rho = |\hat{\mathcal{D}}|/|\mathcal{D}|$.
Given that frame features are pre-extracted, temporal redundancy represents the most prominent overhead in TAS; we thus focus on compression along the temporal axis in this paper.

\subsection{Diffusion and Deterministic DDIM Flow}
Diffusion models~\cite{ho2020ddpm,song2021ddim} are generative models that learn  data distributions by reversing a gradual stochastic corruption process. Given a data sample $x_0$, the forward diffusion process progressively adds Gaussian noise to obtain a noisy latent representation $x_t$ according to:
\begin{equation}
    x_t = \sqrt{\alpha_t}x_0 + \sqrt{1 - \alpha_t} \epsilon,
\end{equation}
where $\epsilon\sim \mathcal{N}(0,I)$ and $\alpha_t$ is a predefined noise schedule controlling the signal-to-noise ratio across timestamps $t$. %

In particular, Denoising Diffusion Implicit Models (DDIM)~\cite{song2021ddim} introduce an efficient sampling formulation by introducing a non-Markovian inference process that supports deterministic trajectories. 
Let $x_T \sim \mathcal{N}(0,I)$ denote Gaussian noise, the inversion and sampling trajectories of DDIM by the forward and reverse flow mappings, respectively:
\begin{equation}
    \Phi_{0\to T} : x_0 \rightarrow \{x_t\}_{t=1}^T, \quad  \text{and} \quad \Phi_{T\to 0} : x_T \rightarrow \{x_t\}_{t=T-1}^0.\label{eq:process}
\end{equation}
The trajectory transition dynamics follow the update rule:
\begin{equation}
    x_{t-1} = \sqrt{\alpha_{t-1}} \hat{x}_0(x_t) + \sqrt{1 - \alpha_{t-1}} \epsilon_\theta (x_t, t),\label{eq:updaterule}
\end{equation}
where $\epsilon_\theta$ is the neural network prediction of the noise component, and $\hat{x_0}(x_t)$ denotes the model's prediction of the clean signal, given by:
\begin{equation}
    \hat{x_0}(x_t) = \frac{x_t - \sqrt{1-\alpha_t}\epsilon_\theta(x_t, t)}{\sqrt{\alpha_t}}.
\end{equation}
In practice, the model parameters are trained using a denoising objective that encourages accurate prediction of the noise component:
\begin{equation}
    \mathcal{L}_{\mathrm{diff}} =\mathbb{E}_{x_0,\epsilon,t}\left[\left|\epsilon - \epsilon_\theta(x_t,t)\right|^2\right],\label{eq:diffloss}
\end{equation}
where $t$ is uniformly sampled from the diffusion timesteps during training.  

Since DDIM admits a non-Markovian formulation, the inversion and sampling trajectories are deterministic once the model parameters and noise schedule are fixed according to~\cref{eq:process,eq:updaterule}. This deterministic property is particularly desirable for dataset condensation, as it enables stable trajectory reconstruction and controlled generation of condensed representations for TAS datasets.

\section{Method}

\subsection{Action Modeling with Diffusion}
For action modeling, we adopt a diffusion-based generative modeling strategy as opposed to the cVAE used in~\cite{ding2025condensing}. Diffusion models are chosen due to their stronger distribution modeling capacity and more flexible latent learning without requiring explicit prior regularization. 

Specifically, we train a diffusion model $\Phi$ to learn the distribution of frame-level features conditioned on the action label and the frame’s relative position within its action sequence, similar to~\cite{ding2025condensing}. For a frame $x_i$ belonging to action $a$ of length $l$ with normalized sequence index $c_i = (i-1)/(\ell-1)$, where $c_i \in [0,1]$, the noise prediction network is therefore parametrized as $\epsilon_\theta(x_{i,t}, a, c_i, t)$ at every step $t$. The action model is trained using the same denoising reconstruction objective as~\cref{eq:diffloss} with additional conditioning variables $a$ and $c$: 
\begin{equation}
        \mathcal{L}_{\mathrm{act}} =\mathbb{E}_{x_i,a,c_i,\epsilon,t}\left[\left|\epsilon - \epsilon_\theta(x_{i,t},a, c_i,t)\right|^2\right],
\end{equation}
where $x_{i,t}$ denotes the latent representation along the DDIM forward flow trajectory $\Phi_{0\to T}(x_i)$ for $x_i$.

\noindent\textbf{Latent Encoding and Generation.} After training, latent representations for any frame can be obtained by directly applying the DDIM forward flow trajectory.
Formally, the latent representation $x_i^*$  is obtained by mapping the frame feature $x_i$ through the trained diffusion model:
\begin{equation}
    x_{i,T}^* = \Phi_{0\to T}(x_i, a ,c_i),\label{eq:latent}
\end{equation}
with $\Phi_{0\to T}(\cdot)$ being the forward process.

Importantly, since DDIM inference adopts a deterministic formulation, obtaining latent representations with high correspondence to the original frame does not require solving an additional optimization problem as in~\cite{ding2025condensing}. Consequently, 
the reconstructed frame feature $\hat{x}_i$ from $x_{i,T}^*$ is expected to closely approximate $x_i$:
\begin{equation}
    \hat{x_i} = \Phi_{T\to 0} (x_{i,T}^*, a, c_i), \quad \text{s.t.},\quad \hat{x}_i\approx x_i,\label{eq:recon}
\end{equation}
where $\Phi_{T\to 0}(\cdot)$ is the reverse flow. 

\begin{figure}[t]          %
  \centering
\begin{overpic}[width=\linewidth, grid=false]{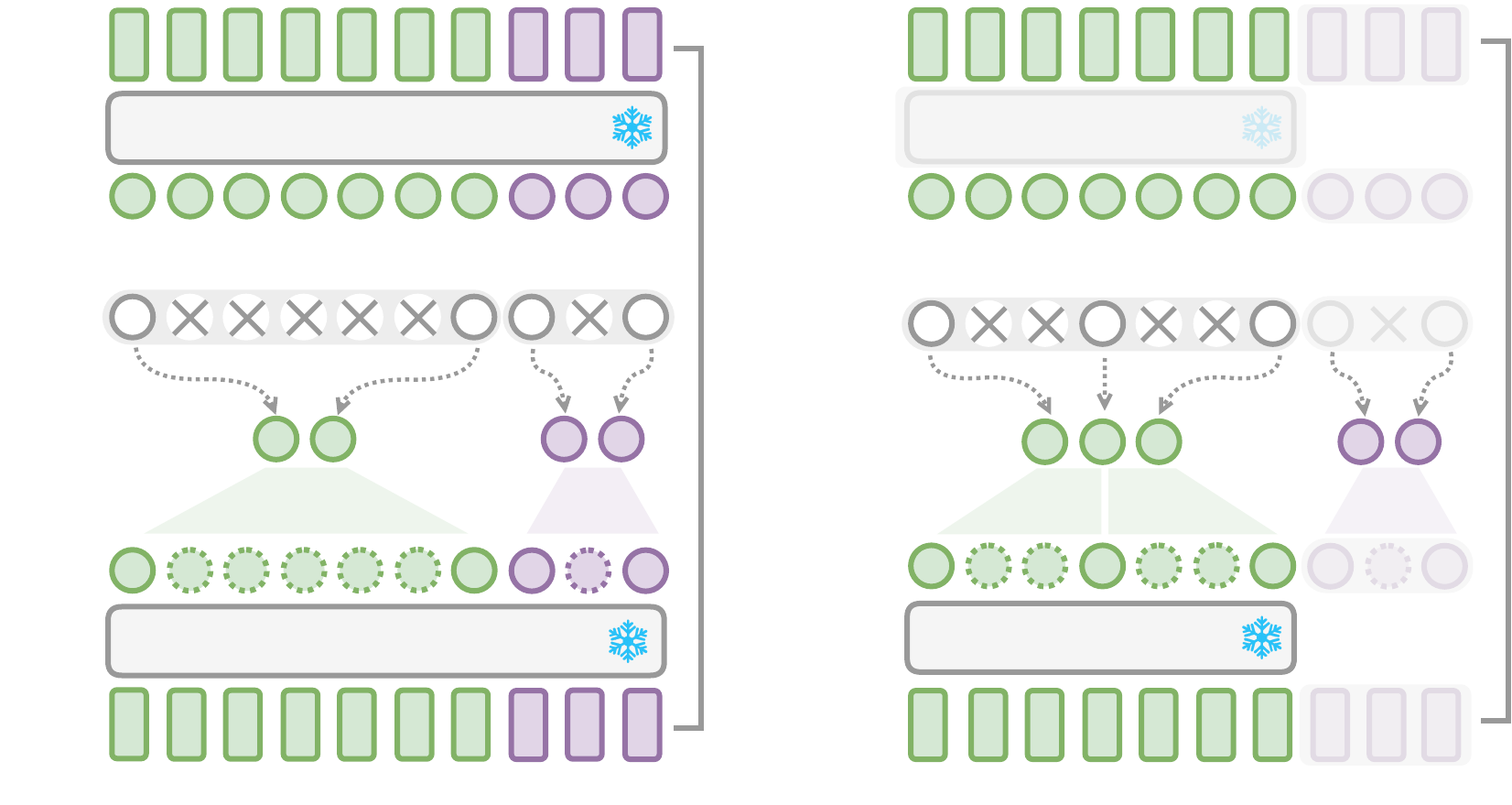}

  \put(1, 6){\tiny recon.}
  \put(2.5, 3.5){\small$\hat{x}_i$}
  
  \put(0, 51){\tiny original}
  \put(2.5, 48.5){\small$x_i$}
  
  \put(1, 41.5){\tiny latent}
  \put(2.5,39){\small$x^*_i$}
  
  \put(1, 33.5){\tiny anchor}
  \put(2.5, 31){\small$i_k$}
  
  \put(1.5, 16.5){\tiny intp.}
  \put(2.5, 14){\small$\hat{x}^*_i$}
  
  \put(0.5, 25){\tiny condense}
  \put(2.5, 22.5){\small$x^*_{i_k}$}

  \put(22, 44){\small$\Phi_{0\to T}$}
  \put(22, 10){\small$\Phi_{T\to 0}$}
  \put(19, 19){\small$\mathcal{I}$}
  \put(38, 19){\small$\mathcal{I}$}
  
  \put(15,0){\scriptsize (a) Anchor Initialization}

  \put(15, 35.5){\footnotesize$\textcolor{action2}{K_1}\!=\!2$}
  \put(34, 35.5){\footnotesize$\textcolor{action3}{K_2}\!=\!2$}
  \put(48, 38.5){\tiny recon error}
  \put(48, 35.5){\footnotesize$\textcolor{action2}{\mathcal{E}_1}>\textcolor{action3}{\mathcal{E}_2}$}

  \put(68, 35.5){\footnotesize$\textcolor{action2}{K_1}\!=\!2\textcolor{red}{+1}$}
  \put(88, 35.5){\footnotesize$\textcolor{action3!40}{K_2}\textcolor{gray!40}{=\!2}$}

  \put(71, 10){\small$\Phi_{T\to 0}$}
  \put(68, 19){\small$\mathcal{I}$}
  \put(76, 19){\small$\mathcal{I}$}
  \put(91, 19){\small\textcolor{gray!40}{$\mathcal{I}$}}
  \put(71, 44){\small\textcolor{gray!40}{$\Phi_{0\to T}$}}
  \put(65,0){\scriptsize (b) Adaptive Anchoring}
\end{overpic}
  \caption{
  Overview of adaptive latent trajectory anchoring. (a) Anchor initialization: Video frames $x_i$ are mapped to latent space $x_i^*$ with the DDIM forward process $\Phi_{0\to T}$. Initial anchors $x^*_{i_k}$ are sampled to represent each segment, which are then interpolated by $\mathcal{I}$ to reconstruct the original frames with $\Phi_{T\to 0}$.  (b) Adaptive Anchoring: The framework dynamically reallocates anchors based on reconstruction error $\mathcal{E}$. Segments with higher error (\eg, $\textcolor{action2}{\mathcal{E}_1}>\textcolor{action3}{\mathcal{E}_2}$) are assigned additional anchors ($\textcolor{red}{+1}$) to capture complex temporal variations, while simpler segments remain intact. This process is then repeated until the anchor budget is exhausted. We omit the conditional variables ($a,c$) from this figure for simplicity.
  }
  \label{fig:overview}
\end{figure}
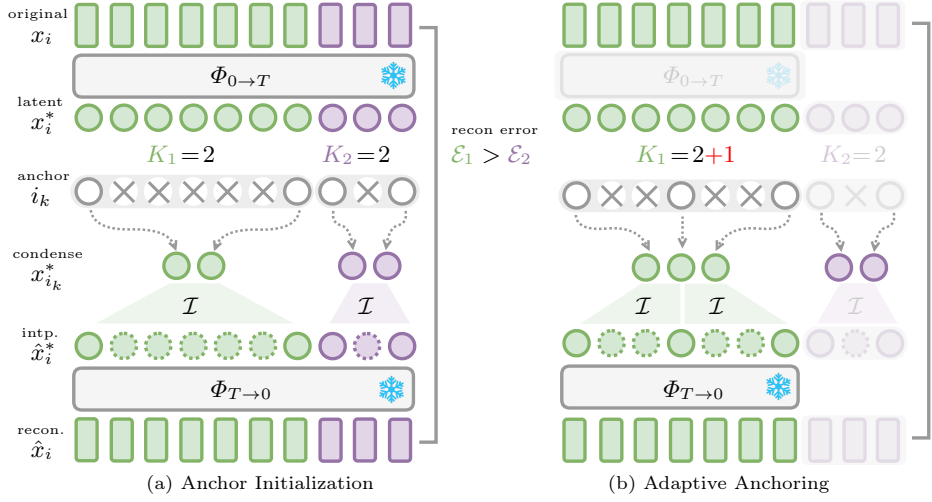

\subsection{Adaptive Latent Trajectory Anchoring}
The primary redundancy in TAS datasets is concentrated along the temporal axis, where consecutive frames within a local neighborhood exhibit high feature affinity. To exploit this, we seek to represent each action segment as a trajectory anchored by a set of sparse latent representations. \cref{fig:overview}  depicts the overall procedure of our adaptive latent trajectory anchoring framework. 

\noindent\textbf{Condensing Actions into Latent Anchors.} 
For a segment $s=\{x_i\}_{i=1}^\ell$ of action class $a$ and duration $\ell$, we identify a compact set of $K$ representative frames $\{x_{i_k}\}_{k=1}^K$, which serve as the anchors for the entire sequence. By storing only these anchors, we reduce the storage footprint of action segments by a factor of $\ell/K$ compared to the original sequence since $K\ll \ell$. A straightforward way to initialize this condensation is to determine the temporal indices $i_k$ for the $k$-th anchor via linear spacing:
\begin{equation}
    i_k = \lceil 1 + \frac{k-1}{K-1}(\ell-1) \rfloor, \quad k= 1,\dots, K,\label{eq:index}
\end{equation}
where $\lceil\cdot\rfloor$ denotes rounding to integer. 
These anchors are then mapped into the latent noise manifold using the DDIM forward process $\Phi_{0\to T}$:
\begin{equation}
   x^*_{i_k} = \Phi_{0\to T}(x_{i_k}, a, c_{i_k}).\label{eq:encode}
\end{equation}
Notably, our diffusion action model enables direct latent mapping via a single forward pass, providing a significant efficiency gain over GNI~\cite{ding2025condensing} that relies on expensive iterative optimization for network inversion.

\noindent\textbf{Latent Temporal Interpolation.} In order to restore temporal resolution during reconstruction for TAS training, we model the action as a continuous latent trajectory to approximate the latent in between the anchors. Specifically, we define an interpolation operator, $\mathcal{I}$, which estimates the latent representations $x^*_j$ for any frames situated between two anchors, \ie, $i_k < j < i_{k+1}$. 
In contrast to the nearest-neighbor ``inflation'' used in~\cite{ding2025condensing}, we have:
\begin{equation}
    x^*_j = \mathcal{I}(x^*_{i_k}, x^*_{i_{k+1}}; \lambda_j)\quad \text{where}\quad \lambda_j = \frac{c_j-c_{i_k}}{c_{i_{k+1}}-c_{i_k}}.\label{eq:interp}
\end{equation}
Interpolating in latent space allows each frame to be assigned a unique latent representation at reconstruction time. This helps to preserve the fine-grained dynamics of action progression in cases where conditioning variables $c$ alone may not suffice. %

\begin{algorithm}[!t]
\small
\caption{Adaptive Latent Trajectory Condensation}
\label{alg:dynamic_anchor}
\begin{algorithmic}[1]

\Require A video $S = \{s_1, \dots, s_N\}$, global anchor budget $B$
\Ensure Condensed anchors $\{K_n\}_{n=1}^N$, %

    \For{each segment $s_n=\{x_i\}_{i=1}^{\ell_n}$} \Comment{Anchor Initialization (\cref{fig:overview}(a))}
    \State Initialize minimal anchors $K_n=2$
        \State Sample anchor indices $\{i_k\}_{k=1}^{K_n}$ \Comment{\cref{eq:index}}
        
        \State Encode anchors $\{x_{i_k}^*\}_{k=1}^{K_n}$ 
        \Comment{\cref{eq:encode}}
        
        \State Interpolate latent trajectory $x_j^*$ 
        \Comment{\cref{eq:interp}}
        
        \State Reconstruct frames $\hat{x}_i$
        \Comment{\cref{eq:recon}}
        
        \State Compute reconstruction error $\mathcal{E}_n$
        \Comment{\cref{eq:difficulty}}
    \EndFor
    \State Build max-priority queue $\mathcal{Q}$ over $\{\mathcal{E}_n\}_{n=1}^N$
    
\While{$\sum_{n=1}^N K_n < B$}    \Comment{Adaptive Anchoring (\cref{fig:overview}(b))}
    \State Select worst segment $n^*$ from the queue
    \Comment{\cref{eq:worst}}
    
    \State Update anchor allocation $K_{n^*}$
    \Comment{\cref{eq:allocate}}

    \State Sample anchors for $s_{n^*}$ with new $K_{n^*}$ 
    \Comment{\cref{eq:index}}

    \State Interpolate, reconstruct and compute error $\mathcal{E}_{n^*}$
    \Comment{\cref{eq:encode,eq:interp,eq:recon,eq:difficulty}}

    \State Insert updated $(n^*, \mathcal{E}_{n^*})$ into $\mathcal{Q}$
    
\EndWhile

\end{algorithmic}
\label{alg:adaptive}
\end{algorithm}

\noindent\textbf{Adaptive Anchor Allocation.} The anchor allocation strategy described above assigns an equal number of anchors to each action segment. While straightforward, enforcing a uniform distribution across segments may be suboptimal at the video level, as action segments can vary substantially in temporal duration and structural complexity. For a video containing $N$ action segments, allocating $K$ anchors per segment yields a total anchor budget of $B = N\times K$. To better utilize this fixed budget, we introduce an adaptive anchor allocation strategy that redistributes anchors across action sequences according to their reconstruction difficulty. 

Let $K_n$ denote the number of anchors assigned to segment $s_n$, we initialize each action segment with a minimal interpolation capacity $K_n = 2$, which ensures that latent interpolation between anchors is well-defined. To guide adaptive allocation, we measure the reconstruction difficulty of each segment using the mean reconstruction error:
\begin{equation}
    \mathcal{E}_n = \mathbb{E}_{x_i\in s_n}[|x_i-\Phi_{T\to 0}(x^*_i,a,c_i)|^2]\label{eq:difficulty}
\end{equation}
where $x^*_i$ denotes the interpolated latent representations obtained by~\cref{eq:interp} under the current anchor configuration. Reconstruction objectives are widely used in representation learning as a proxy to indicate that the underlying structure of the data is well captured~\cite{he2022masked}. Here, we interpret the reconstruction residual $\mathcal{E}_n$ as a measure of how well the current set of anchors capture the temporal dynamics of the sequence. 

At each iteration, an additional anchor is assigned to the segment with the largest reconstruction error, \ie,:
\begin{equation}
    n^* =\argmax_n \mathcal{E}_n,\label{eq:worst}
\end{equation}
followed by updating 
\begin{equation}
    K_{n^*}\leftarrow K_{n^*}+1.\label{eq:allocate}
\end{equation}
After each update, anchor positions for segment $s_{n^*}$ are re-sampled via~\cref{eq:index} under the updated $K_{n^*}$, and the corresponding latent representations are recomputed.
The interpolation operator is then re-applied to obtain updated latent approximations for that segment before the next allocation step.

This iterative procedure continues until the global anchor budget $B$ is exhausted. By allocating anchors according to segment-wise reconstruction difficulty, the proposed strategy adaptively matches representation capacity to segment-wise reconstruction difficulty, assigning more anchors to segments that require finer temporal modeling. The overall adaptive anchor allocation is summarized in~\cref{alg:adaptive}.

\subsection{Computational Complexity}
Our framework achieves significantly lower latency by leveraging the deterministic flow of DDIM. For a video of $N$ segments, our standard anchoring strategy of inversion and reconstruction process with $T$ diffusion timesteps requires: $\mathcal{O}(NT)$. 
For the adaptive variant, anchoring is repeated until the anchor budget $B=N\times \bar{K}$ is exhausted, yielding $\mathcal{O}(N\bar{K}T)$, where $\bar{K}$ is the average number of anchors per segment. 
While with GNI~\cite{ding2025condensing}, for the same video, the latent codes are optimized via $S$ optimization steps, yielding a total complexity of $\mathcal{O}(NS)$.

In practice, $T$ is typically set to small values (\eg, 10 or 50), whereas GNI~\cite{ding2025condensing} requires thousands of iterations ($S\!=\!10,000$) of optimization. This ensures that even with the iterative nature of adaptive anchor allocation, the cumulative cost $N\bar{K}T$ remains orders of magnitude smaller than the $\mathcal{O}(NS)$ cost of GNI while achieving greater temporal expressiveness through latent anchoring.

\subsection{Decoding for TAS Training}
The condensed latent representation obtained from the adaptive trajectory condensation process is used to construct a compact surrogate dataset for TAS training. After condensation, each action segment is represented by a sparse set of anchor latent codes $\{x^*_{i_k}\}_{k=1}^K$. Formally, the reconstructed frame-level representation of a video sequence is given by:
\begin{equation}
    \hat{x}_i = \Phi_{T\to 0} (\mathcal{I}(x_{i_k}^*,x_{i_{k+1}}^*;\lambda_i), a, c_i).
\end{equation}

The TAS model is then trained on the reconstructed dataset with the standard loss functions introduced in~\cref{eq:tas} by substituting $x$ with $\hat{x}$:
\begin{equation}
    \mathcal{L}_\text{tas} = \mathcal{L}_\text{cls} (\hat{x},y)+ \lambda \mathcal{L}_\text{sm} (\hat{x}).
\end{equation}

\section{Experiment}
\subsection{Datasets and Evaluation Metrics}
\textbf{Datasets.}
We evaluate our method on three widely used TAS benchmarks that vary in scale and storage requirements: GTEA~\cite{fathi2011learning}, 50Salads~\cite{stein2013combining}, and Breakfast~\cite{kuehne2014language}. 
\textbf{GTEA}~\cite{fathi2011learning} consists of 28 egocentric kitchen videos spanning 7 high-level activities and 11 action classes. The videos are relatively short and exhibit limited intra-class variation. 
\textbf{50Salads}~\cite{stein2013combining} contains 50 long videos of salad preparation annotated with 19 action classes. It features extended temporal durations and more complex action transitions.  
\textbf{Breakfast}~\cite{kuehne2014language} is a large-scale benchmark comprising 1,712 videos covering 10 breakfast preparation activities and 48 fine-grained action classes. Each video contains 5 to 14 action segments on average, with substantial variability in duration and ordering. 

For all datasets, we use pre-extracted I3D features~\cite{carreira2017quo} and follow the standard evaluation splits. While I3D compresses spatial information by mapping RGB frames into a feature space, the original temporal resolution of the video is preserved.

\noindent\textbf{Evaluation Metrics.}
We follow the standard TAS evaluation protocol and report three metrics: frame-wise accuracy (Acc), segmental edit score (Edit), and F1 score at overlap thresholds of 10\%, 25\%, and 50\%.

\subsection{Implementation}
Our diffusion model was implemented as an $\epsilon$-predictor with a lightweight architecture design. The timestep is encoded using a single linear layer, then concatenated with the other inputs and passed through a two-layer linear module with a bottleneck compression ratio of 16. We used a learning rate of 0.001 and trained the model for 2K epochs. We set the average anchor number per segment to be $\bar{K}=8$.

\noindent\textbf{Baselines.}
Following~\cite{ding2025condensing}, we implement the following baselines for comparison:

\noindent -- \textbf{Mean:} A simple condensation baseline where frame features are averaged within each action segment. The averaged feature is then temporally expanded during reconstruction to match the original segment length.

\noindent -- \textbf{Intp.:} This method stores the first and last frame of each action segment and reconstructs intermediate frames through linear interpolation between them.

\noindent -- \textbf{Coreset:} This approach uses herding to select the frame feature closest to the mean feature of each segment. The selected features are then temporally upsampled to restore the original temporal resolution.

\noindent --  \textbf{GNI~\cite{ding2025condensing}:} This method employs a time coherent VAE. Each action sequence is divided into temporal chunks, and a common latent representation is optimized for each chunk. During reconstruction, each latent is expanded to the corresponding chunk length.
We optimize latent codes for 10K steps.
    
\noindent --  \textbf{Original:} This is the standard setup where full original frame features are used, which we consider as the upper bound.

\begin{table*}[t]
\centering

\resizebox{\textwidth}{!}{
\begin{tabular}{lcccccccccc}
\toprule
& & \multicolumn{3}{c}{{GTEA (256MB)}}
 & \multicolumn{3}{c}{{50Salads (4.7GB)}}
 & \multicolumn{3}{c}{{Breakfast (27.4GB)}} \\
\cmidrule(lr){2-2}\cmidrule(lr){3-5} \cmidrule(lr){6-8} \cmidrule(l){9-11}
&$\Bar{K}$ & Acc & Edit & F1@10/25/50 
 & Acc & Edit & F1@10/25/50
 & Acc & Edit & F1@10/25/50  \\
\midrule

\multicolumn{11}{c}{{MSTCN}~\cite{farha2019ms}} \\
\midrule

Mean
&1& 69.4 & 67.5 & 72.6/68.6/52.8
& 70.0 & 46.7 & 54.2/49.5/40.2
& 48.0 & 33.2 & 29.6/25.5/17.6\\

Intp.
&2& 61.1 & 69.6 & 74.6/64.4/46.8
& 49.9 & 53.9 & 54.4/49.6/34.6
& 33.6 & 42.9 & 38.7/31.8/19.3\\

Coreset
&1& 60.3 & 63.4 & 65.4/59.2/44.2
& 69.6 & 62.5 & 65.7/63.1/56.6
& 50.2 & 41.3 & 36.7/31.6/22.3 \\

GNI~\cite{ding2025condensing}
&8& 66.7 & 71.9 & 73.9/68.9/45.9
& \textbf{73.5} & 43.8 & 49.0/46.6/38.6
& 37.9 & 44.5 & 38.1/33.5/24.7 \\

GNI~\cite{ding2025condensing}
&64& 66.6 & 66.9 & 74.2/69.9/48.8
&72.8 & 44.2 & 49.3/45.0/40.2
&47.8 & 54.0 & 46.9/41.9/30.9\\\midrule

\rowcolor{gray!20}Ours$^\dagger$
&8& 71.2 & \textbf{80.6}& \textbf{86.5}/\textbf{80.5}/58.6
& 72.7 & \textbf{65.1} & \textbf{71.9}/\textbf{68.6}/\textbf{57.9}
& 54.8 & \textbf{67.2} & \textbf{63.8}/57.6/45.1\\

\rowcolor{gray!20}Ours
&8& \textbf{72.1} & 79.1 & 85.9/78.0/\textbf{61.4}
& 70.1 & 62.3 & 68.8/65.3/53.5
&\textbf{ 63.4} & 65.6 & \textbf{63.8}/\textbf{58.4}/\textbf{46.1} \\\midrule

 Original 
&$\ell$& 72.7 & 73.8 & 79.9/73.7/59.4
& 74.6 & 60.0 & 66.9/64.7/56.2
& 67.9 & 67.9 & 67.7/61.8/49.5\\

\midrule
\multicolumn{11}{c}{{ASFormer}~\cite{yi2021asformer}} \\
\midrule

Mean &1
& 70.7 & 73.3 & 78.2/76.9/67.3
& 62.2 & 40.7 & 47.9/42.2/33.4
& 51.8 & 46.5 & 44.9/39.8/28.1\\

Intp. & 2
& 64.1 & 72.7 & 76.5/71.9/54.9
& 54.8 & 48.2 & 55.4/49.5/31.9
& 43.1 & 48.3 & 47.3/40.1/25.6\\

Coreset & 1
& 67.2 & 69.4 & 73.6/72.3/55.3
& 66.8 & 45.3 & 53.6/48.4/38.2
& 49.9 & 51.0 & 47.1/41.6/30.2\\

GNI~\cite{ding2025condensing} & 8
& 71.6 & 76.4 & 80.8/79.4/63.4
& 70.5 & 51.4 & 61.6/57.6/47.3 
& 62.8 & 66.3 & 65.8/59.4/47.2 \\

GNI~\cite{ding2025condensing} & 64
& 70.8 & 73.5 & 81.2/77.8/62.1
& 72.1 & 49.9 & 59.7/56.1/46.2
& 61.1 & 64.6 & 63.5/58.4/45.9\\\midrule

\rowcolor{gray!20}Ours$^\dagger$& 8
& \textbf{73.7} & \textbf{81.5} & 84.3/81.5/71.8
& \textbf{75.1} & \textbf{64.9} & \textbf{71.7}/\textbf{68.8}/\textbf{57.6}
& 65.3 & \textbf{71.5} & 70.1/65.2/48.4  \\

\rowcolor{gray!20}Ours & 8
& 73.4 & 80.5 & \textbf{85.2}/\textbf{83.1}/\textbf{73.9}
& 71.2 & 60.7 & 69.2/64.9/54.4 
& \textbf{68.0} & 71.3 & \textbf{71.4}/\textbf{66.3}/\textbf{52.9}  \\\midrule

Original & $\ell$
& 74.3 & 77.4 & 82.9/80.1/73.1
& 77.1 & 67.6 & 77.1/73.1/61.7
& 70.7 & 72.0 & 73.6/69.0/56.1\\
\bottomrule
\end{tabular}}
\caption{Comparison of dataset condensation performances on three TAS benchmarks using MSTCN and ASFormer backbones. $^\dagger$ indicates fixed anchor allocation. Our trajectory-based condensation variations consistently outperform existing baselines and shows competitive performance when training with original data. %
}
\label{tab:effectiveness}
\end{table*}

\subsection{Effectiveness}
\cref{tab:effectiveness} compares different strategies for TAS condensation across common benchmarks. Since GNI additionally applies an 8$\times$ compression along the feature dimension, we reproduce and report GNI under two settings: GNI with $\bar{K}=8$, which matches our temporal compression factor $\bar{K}$, and GNI with $\bar{K}=64$, which matches our overall compression ratio $\rho$. Our method is evaluated in two variants: the primary adaptive latent trajectory anchoring (Ours) and a variant using standard anchor allocation (Ours$^\dagger$).
As we can see, both of our variants surpass the strongest baseline GNI~\cite{ding2025condensing} by significant margins across all metrics. For instance, on Breakfast with the ASFormer backbone, our primary variant achieves 68.0\% accuracy, representing a 5.2\% absolute improvement over GNI (62.8\%) and an 18.1\% improvement over the Coreset baseline (49.9\%).
The two variants also exhibit complementary strengths. The standard version (Ours$^\dagger$) can achieve higher Acc in some cases, such as 75.1\% accuracy on 50Salads. In contrast, the adaptive variant (Ours) shows better long-term temporal consistency, reflected by higher values on segmental metrics. This suggests that the adaptive strategy is effective in modeling the action evolution of long and complex actions where more anchors are allocated.

Last but not least, our method remarkably narrows the performance gap relative to the original full dataset. For example, on 50Salads with ASFormer, our method reaches 75.1\% accuracy, coming within 2\% of the performance achieved using the full dataset, while using only 1.4\% of storage.

\subsection{Ablation Study and Analysis}
\noindent \textbf{Anchor budget.} Table~\ref{tab:budget} reports the effect of the mean anchor budget $\bar{K}$ on condensation performance. As we can see, increasing $\bar{K}$ tends to improve the segmentation performance across all metrics, reflecting the benefit of denser temporal coverage. However, returns diminish beyond $\bar{K}=8$: increasing to $\bar{K}=10$ yields only a marginal gain in accuracy ($72.1\% \rightarrow 73.3\%$) at the cost of a $21\%$ increase in storage ($\rho: 0.24 \rightarrow 0.29$). We therefore set $\bar{K}=8$ as our default. 

On the other hand, our adaptive anchoring strategy (Ours) generally outperforms the uniform baseline (Ours$^\dagger$) across various budget constraints ($\bar{K}$). This performance gain is particularly evident at lower anchor budgets, where the adaptive mechanism adaptively redistributes resources to segments that are most challenging to reconstruct.

\begin{table*}[t]
\centering

\begin{tabular}{lccccccccc}

\toprule

&  & &\multicolumn{3}{c}{{Ours$^\dagger$}} 
& &\multicolumn{3}{c}{{Ours}} \\\cmidrule(lr){4-6}\cmidrule(lr){8-10}
$\bar{K}$ & $\rho$ & & Acc & Edit & F1@10/25/50
& & Acc & Edit & F1@10/25/50  \\

\midrule

2
& 0.06 & & 64.4 & 79.5 & 78.2/68.4/51.1
& & - & - & - / - / - \\

4
& 0.12 & &  68.6 & \textbf{81.9} & 84.8/76.5/57.6
& &70.5 & 78.8 & 81.4/72.8/59.3 \\

6
& 0.18 & & 71.2 & 79.4 & 85.9/76.3/57.8
& &72.5 & \textbf{80.0} & 84.8/76.8/58.7 \\

\rowcolor{gray!20}
8
& 0.24 & & 71.2 & 80.6 & \textbf{86.5}/\textbf{80.5}/58.6
& &72.1 & 79.1 & \textbf{85.9}/\textbf{78.0}/61.4 \\

10
& 0.29 & & \textbf{73.3} &  78.4 & 84.7/78.8/\textbf{63.5}
& &\textbf{73.8} & 77.2 & 83.3/75.5/\textbf{61.5}\\

12
& 0.32 & & 71.5 & 81.2 & 82.8/75.5/61.5
& &71.8 & 75.5 & 82.4/76.7/60.2 \\

\bottomrule

\end{tabular}

\caption{Performance comparison on GTEA dataset under MSTCN backbone of our methods with different anchor budget.}
\label{tab:budget}
\end{table*}

\begin{table}[t]
    \centering
    \begin{minipage}[t]{0.48\textwidth}
        \vspace{0pt} %
        \centering
        \small
        \begin{tabular}{lcccccc}
            \toprule
            & $\bar{K}$ & $\rho$ & Acc & Edit & F1@10/25/50 \\
            \midrule
            \multirow{2}{*}{GNI~\cite{ding2025condensing}} & 8 & 0.03 & 66.7 & 71.9 & 73.9/68.9/45.9 \\
            & 64 & 0.24 & 66.6 & 66.9 & 74.2/69.9/48.8 \\
            \midrule
            Ours$^\dagger$ & 8 & 0.24 & 71.2 & 80.6 & 86.5/80.5/58.6 \\
            Ours & 8 & 0.24 & 72.1 & 79.1 & 85.9/78.0/61.4 \\
            \bottomrule
        \end{tabular}
        \caption{The expressiveness comparison of action models on GTEA. Increasing the number of latent codes for GNI leads to performance plateau, likely due to an inherent information bottleneck. While our approach achieves significant boost with same compression ratio.}
        \label{tab:model}
    \end{minipage}
    \hfill
    \begin{minipage}[t]{0.48\textwidth}
        \vspace{0pt} %
        \centering
		\includegraphics{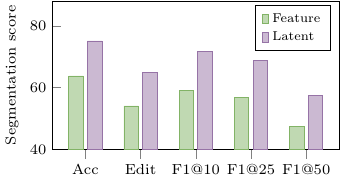}
        \captionof{figure}{Comparison between interpolation in the feature space and interpolation in the latent space on 50salads. Latent interpolation shows consistent performance gain over direct feature interpolation.}
        \label{fig:space}
    \end{minipage}
\end{table}

\begin{figure}[h!]
    \centering
    \hspace{-1em}
    \includegraphics{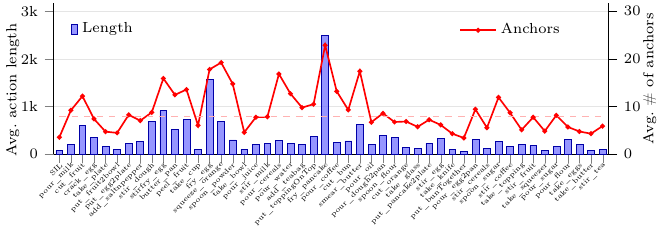}
\caption{Analysis of dynamic anchor allocation on Breakfast. Actions are ordered by decreasing frequency. Longer actions generally receive more anchors.}\label{fig:len_anchor}
\end{figure}

\noindent \textbf{Expressiveness of action models.}
The results reported in~\cref{tab:model} provide an empirical investigation into the representational capacity of different action models. A critical observation for the GNI baseline is that increasing the number of sub-segments ($\bar{K}$) from 8 to 64, the model exhibits a clear performance plateau, while Edit score degrades from 71.9\% to 66.9\%. A similar trend is also reported by~\cite{ding2025condensing} suggesting that the expressiveness of the underlying VAE framework is inherently bounded. In contrast, our proposed method consistently outperforms them significantly with the same compression ratio $\rho$.  

\noindent \textbf{Interpolation space.} 
\cref{fig:space} compares interpolation in feature space and latent space on 50Salads dataset. Overall, latent space interpolation consistently yields better segmentation performance across all metrics, suggesting that trajectory modeling in the latent representation better preserves temporal semantic structure than direct interpolation in raw feature space. This supports our design choice of performing condensation through anchor interpolation in a learned latent space.

\noindent \textbf{Dynamic Anchor Distribution Analysis.} We next study how these dynamic anchors are distributed across action segments. \cref{fig:len_anchor} shows the average number of anchors assigned to each action category under the dynamic anchor allocation setting. While actions are ordered by frequency, no clear correlation between frequency and anchor count is observed. Instead, categories with longer average durations tend to receive more anchors, suggesting that the allocation adapts primarily to temporal scale. This adaptive behavior enables the model to provide finer duration coverage for long actions while avoiding unnecessary anchors for short actions.

\begin{figure}[!t]
\centering
\begin{overpic}[width=\linewidth, grid=false]{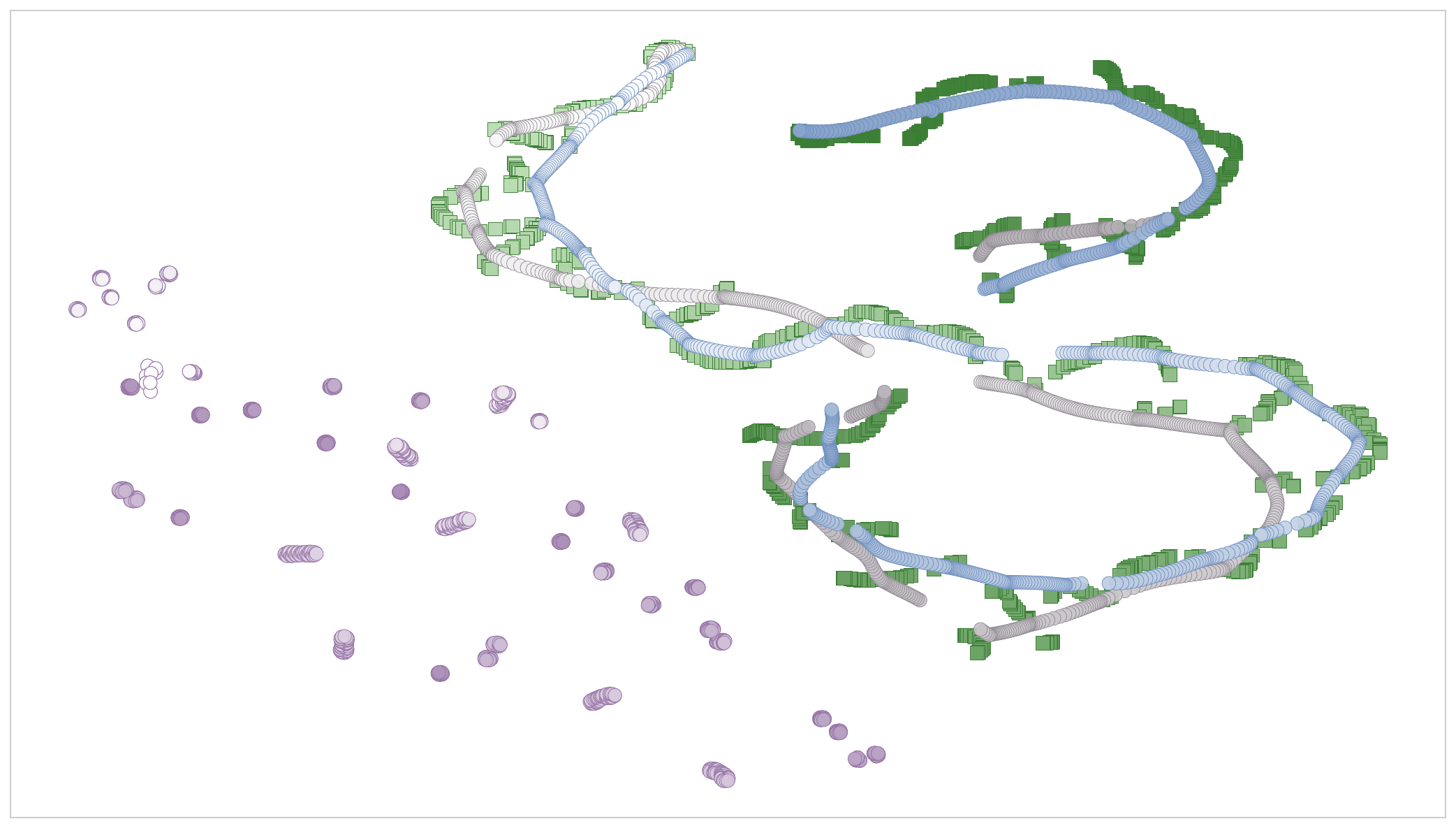}
    \put(85.5, 2){%
      \includegraphics{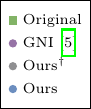}
    }
\end{overpic}
\caption{
Comparative t-SNE visualizations of temporal feature trajectories for action \texttt{take\_bowl} in video \texttt{P03\_cam01\_P03\_cereals} from the Breakfast dataset. We evaluate the alignment between the Original features (\textcolor[rgb]{0.51,0.70,0.40}{green squares}) and three models: GNI~\cite{ding2025condensing} (\textcolor[rgb]{0.59,0.45,0.65}{purple}) , Ours$^\dagger$ with fixed anchors (\textcolor[rgb]{0.56,0.53,0.57}{gray}), and Ours with adaptive anchors (\textcolor[rgb]{0.42,0.56,0.75}{blue}). Variation in color intensity indicates temporal progression. Our adaptive approach produces feature trajectories that most closely approximate the original. (Best viewed when zoomed in.)
}
\label{fig:visualization}
\end{figure}

\noindent \textbf{Visualization.} 
To assess the fidelity of the condensed representations, we visualize the temporal feature trajectories using t-SNE~\cite{van2008visualizing} for a representative sequence from the Breakfast dataset in~\cref{fig:visualization}. As we can see, the optimization-based GNI~\cite{ding2025condensing} produces fragmented clusters that fail to capture the sequential manifold of the original features. While our method with a fixed budget recovers the global flow, it tends to over-simplify intricate temporal variations in complex feature regions. In contrast, our full adaptive framework achieves superior alignment by dynamically concentrating anchors where the trajectory is most volatile.
Our diffusion-based approach faithfully preserves both local dynamics and global progression, as evidenced by the reconstructed trajectory's high proximity to the ground truth.

\section{Conclusion}
In this paper, we present a diffusion-based TAS condensation framework that replaces expensive iterative optimization with deterministic latent trajectory anchoring. By leveraging the bijective properties of DDIM, we represent action segments as trajectories in the latent space, preserving fine-grained temporal dynamics that traditional cVAE-based methods tend to over-smooth.
In addition, our adaptive budgeting strategy ensures high-fidelity reconstructions by concentrating representation capacity on complex segments. Results across common TAS benchmarks demonstrate that our approach achieves competitive performance to full real data training while using as little as 1.4\% of the original storage. This shift toward optimization-free, trajectory-based compression offers a scalable and efficient path for managing large-scale video data.

\noindent\textbf{Acknowledgment.} This research / project is supported by the Ministry of Education, Singapore, under the Academic Research Fund Tier 1 (FY2025).

\bibliographystyle{splncs04}
\bibliography{main}
\end{document}